\pdfoutput=1

\documentclass[11pt]{article}

\usepackage[preprint]{acl}
\usepackage{times}
\usepackage{latexsym}
\usepackage[normalem]{ulem}  
\renewcommand{\emph}[1]{\uline{#1}} 
\definecolor{darkgreen}{RGB}{0,100,0}
\usepackage[T1]{fontenc}

\usepackage[utf8]{inputenc}

\usepackage{microtype}

\usepackage{inconsolata}

\usepackage{graphicx}
\usepackage{amsmath}
\usepackage{amssymb}
\usepackage{booktabs}
\usepackage{algorithmic}
\usepackage{amsfonts}
\usepackage{subfigure}
\usepackage{url}
\usepackage{multirow}
\usepackage{multicol}
\usepackage{arydshln}
\usepackage{bbm}
\usepackage[ruled,vlined]{algorithm2e}
\usepackage{colortbl}
\usepackage[most]{tcolorbox}
\tcbuselibrary{skins, breakable}

\usepackage{dsfont}

\newcommand{\ModelName}{\textit{SymbolicThought}}

\title{\textit{SymbolicThought}: Integrating Language Models and Symbolic Reasoning for Consistent and Interpretable Human Relationship Understanding}

\author{Runcong Zhao$^{1*}$, Qinglin Zhu$^{1*}$,  Hainiu Xu$^1$, Bin Liang$^2$$ \textbf{, Lin Gui$^1$}$, Yulan He$^{1,3}$\\
  $^1$King's College London,  $^2$MoE Lab, CUHK, $^3$The Alan Turing Institute\\
  \texttt{\{runcong.zhao, qinglin.1.zhu, hainiu.xu\}@kcl.ac.uk, } \texttt{bin.liang@cuhk.edu.hk}\\
  \texttt{\{lin.1.gui, yulan.he\}@kcl.ac.uk} }

\begin{document}
\maketitle
\def\thefootnote{*}\footnotetext{Equal contribution.}\def\thefootnote{\arabic{footnote}}
\begin{abstract}
Understanding character relationships is essential for interpreting complex narratives and conducting socially grounded AI research. However, manual annotation is time-consuming and low in coverage, while large language models (LLMs) often produce hallucinated or logically inconsistent outputs. We present \textit{SymbolicThought}, a human-in-the-loop framework that combines LLM-based extraction with symbolic reasoning. The system constructs editable character relationship graphs, refines them using seven types of logical constraints, and enables real-time validation and conflict resolution through an interactive interface. To support logical supervision and explainable social analysis, we release a dataset of 160 interpersonal relationships with corresponding logical structures. Experiments show that \textit{SymbolicThought} improves annotation accuracy and consistency while significantly reducing time cost, offering a practical tool for narrative understanding, explainable AI, and LLM evaluation. A screencast\footnote{\url{https://youtu.be/GmtI4qXumqI}} and demo\footnote{\url{http://113.45.152.150:8888/}} are available online.

\end{abstract}

\section{Introduction}

Recent advancements in artificial intelligence, particularly the emergence of large language models (LLMs), have reignited discussions around the potential for artificial general intelligence. The increasingly human-like capabilities of AI are also drawing significant attention from social science researchers \cite{DBLP:journals/ipm/XuSRGPLSH24}. One of the most critical functions in social science research is understanding the relationships between individuals. This understanding is foundational to various disciplines, including journalism, legal document analysis, historical studies, and literary interpretation.

Many existing studies suggest that social relationships can be modeled as compact graphs, also known as small-world networks. These graphs are characterised by a high clustering coefficient and short path lengths. Consequently, such networks typically contain \textbf{small but densely connected subgraphs} (communities), where nodes within a community are highly likely to be connected. Connections between different communities are often \textbf{facilitated by a few key nodes} (individuals).

However, constructing such graphs solely from plain text descriptions presents a significant challenge. This is because the dense local structure of social communities requires a method capable of capturing most intra-community relationships, while the accurate identification of key inter-community connections demands high precision in detecting critical relationships. Traditional methods rely on either \textbf{crowdsourced annotations} or automated approaches based on LLMs. Both methods exhibit substantial limitations.

On one hand, \textbf{human annotation} is generally accurate but demands significant cognitive effort and time investment, which usually lead to \textbf{low coverage} in the annotation. In particular, essential cross-sentence and long-distance connections are frequently missed or inaccurately recorded \citep{conan-zhao2024, sancheti-rudinger-2025-tracking}.
On the other hand, \textbf{LLM-based methods} provide scalability and efficiency but suffer from issues such as hallucinations, semantic inconsistency, and fragmented narrative representations \citep{liu2023lost-in-the-middle, longdoc-coreference-guo2023, West2023AIparadox, yuan-etal-2024-evaluating}, which often fail to predict key relations precisely \citep{DBLP:journals/corr/abs-2401-06080}. 

To address these challenges, we propose \textbf{\ModelName}, a hybrid framework designed for extracting character relationship graphs from narrative texts, combining the scalability and reasoning ability of LLMs with the precision and interpretability provided by human annotators. {\ModelName}~integrates an LLM-based extraction mechanism with \textbf{interactive human-in-the-loop verification} supported by a novel \textbf{symbolic reasoning module} with an editable social knowledge base. Crucially, \ModelName~leverages an intuitive visual interface that highlights relevant textual evidence and systematically detects logical inconsistencies or conflicts, significantly enhancing annotators' focus and efficiency. 
By automatically surfacing critical information and reducing annotators’ cognitive load in searching through extensive narratives, our system empowers humans to direct their attention toward complex judgments and nuanced decisions, which are tasks that current fully automated systems are still unable to handle.
Through this structured, human-centered annotation pipeline, \ModelName~provides a solution that is:  
\begin{itemize}
    \item \textbf{Interpretable}, where the symbolic reasoning driven by the LLM is easily traceable, localisable, and understandable by users;
    \item \textbf{Extendable}, where users can edit the rules underlying the symbolic reasoning;
    \item \textbf{Complete}, where the automatic annotation framework is designed to generate all possible relation graphs for a given character list;
\end{itemize}

\noindent This enables the construction of an accurate and comprehensive static character relationship graphs, laying a strong foundation for advanced narrative analysis and socially aware AI tools across the humanities and social sciences.
\section{Architecture of \ModelName}

Given a narrative document $\mathcal{D}$, the goal of \ModelName~is to construct a character relationship graph $\mathcal{G} = (\mathcal{V}, \mathcal{E})$, where $\mathcal{V}$ is the set of unique, verified character nodes and $\mathcal{E}$ is the set of semantically and logically consistent relationships between them.
The pipeline consists of two main stages: (1) \textbf{Character Extraction:} An LLM-based module $\mathcal{F}{\text{char}}$ proposes candidate entities from $\mathcal{D}$, which are then verified, grouped, or corrected by annotators to yield the finalised node set $\mathcal{V}$.
(2) \textbf{Relationship Extraction:} A downstream extraction module $\mathcal{F}{\text{ext}}$ predicts pairwise relationships conditioned on both $\mathcal{D}$ and $\mathcal{V}$. These predictions are refined by a symbolic module $\mathcal{F}_{\text{ref}}$ and further reviewed by annotators to produce the final edge set $\mathcal{E}$.

\begin{figure*}[th!]
    \centering
    \includegraphics[width=0.95\linewidth]{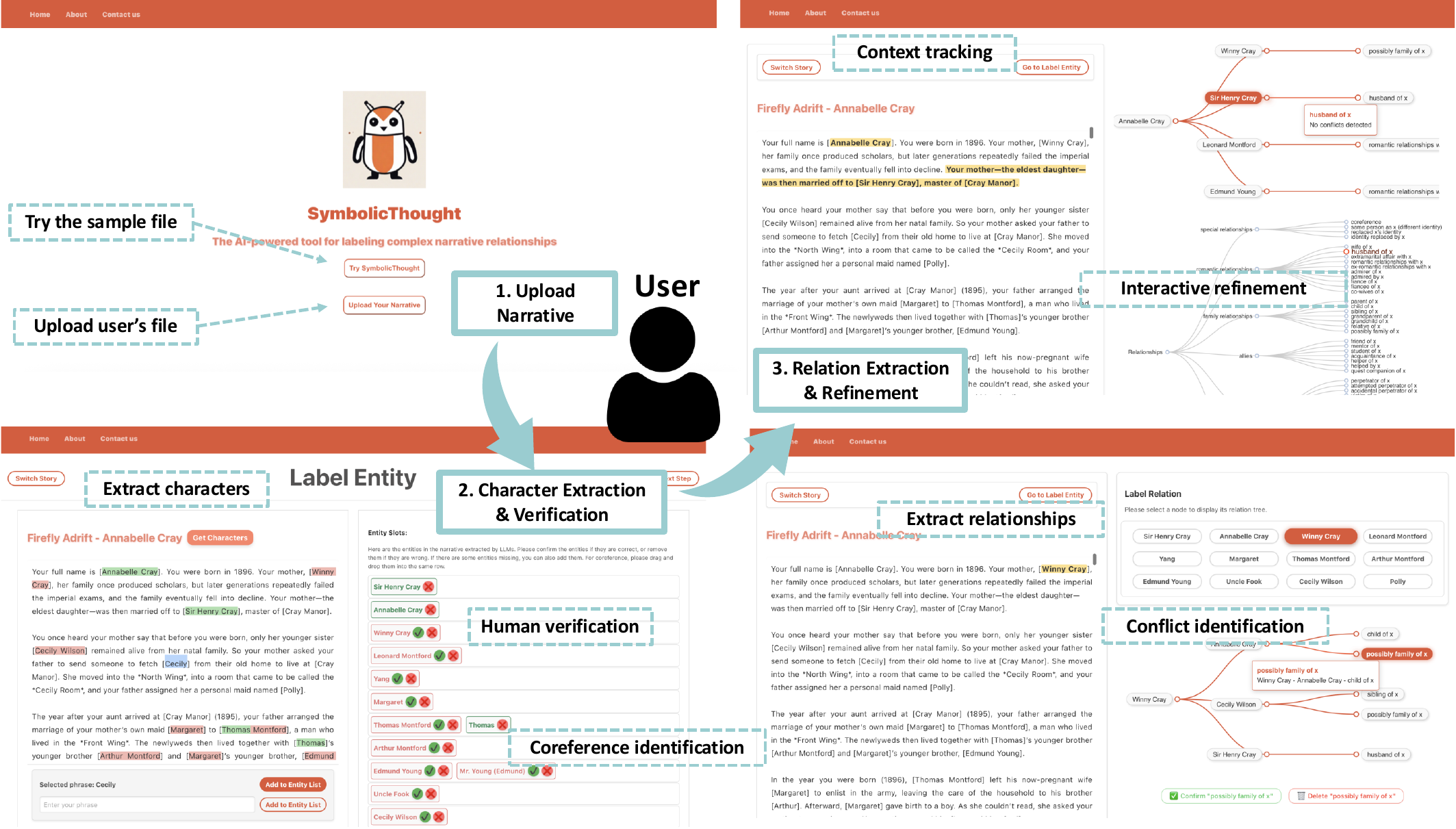}
    \caption{The \textit{SymbolicThought} annotation workflow. Users (1) upload a narrative document, (2) verify or modify automatically extracted character mentions, and (3) confirm, edit, or refine LLM-generated relationships with real-time symbolic consistency checks. The interface provides visual feedback, alerts for logical conflicts, and suggestions for inferable relations, supporting accurate and efficient human-in-the-loop annotation.}
    \label{fig:workflow}
\end{figure*}


\subsection{Character Extraction}
This step corresponds to the lower-left region of Figure~\ref{fig:workflow}. The goal is to identify a complete and clean set of character entities $\mathcal{V}$ from the input narrative document $\mathcal{D}$, which serves as the foundation for downstream relation extraction.

\paragraph{Explorer: Node Candidate Generation.}
We run the character extractor $\mathcal{F}_{\text{char}}$ on the corpus $\mathcal{D}$ independently $n_c$ times, each with temperature sampling to broaden coverage.  Mentions that recur in at least $\tau_c$ runs are retained, yielding a consensus set of candidate nodes:
\[
\mathcal{V}_{\text{raw}}
=\left\{c\;\middle|\;
  \sum_{i=1}^{n_c}
  \mathbb{I}\!\bigl[
    c\in\mathcal{F}_{\text{char}}^{(i)}(\mathcal{D})
  \bigr]\ge\tau_c
\right\}.
\]
This frequency filter maximises recall while suppressing spurious extractions, providing a high-coverage node list for subsequent verification.

\paragraph{Verifier: Human Verification.}
$\mathcal{V}_{\text{raw}}$ is displayed in a front-end interface.
Annotators can accept or reject suggestions, highlight new spans, or add missing entries. Mentions \emph{shift} from \textcolor{red}{red} (unconfirmed) to \textcolor{darkgreen}{green} (confirmed), providing instant feedback.
To improve coherence, our interface supports two editing operations: (1) \emph{Alias Resolution:} Many characters appear under multiple aliases (e.g., \textit{``Elizabeth'', ``Her Majesty'', or ``Queen Elizabeth II''}). Annotators can merge these into a single canonical entity via drag-and-drop operations.
(2) \emph{Entity Disambiguation:} In contrast, distinct characters may share the same surface name (e.g., \textit{``Alexandre Dumas'' may refer to either the father or the son}). Annotators can split such ambiguous mentions into separate rows and assign more specific names, such as \textit{``Alexandre Dumas Sr.'' and ``Alexandre Dumas Jr.''}.

\subsection{Relation Extraction}
This stage corresponds to the right-hand pane of Figure~\ref{fig:workflow}. Its goal is to transform the narrative’s candidate relations $\mathcal{E}$ into a directed graph that captures social, familial, and other types of interactions among the characters, thereby providing a structured representation for subsequent analysis.

\paragraph{Explorer: Edge Candidate Generation.}
With the character inventory fixed, we run the relation extractor $\mathcal{F}_{\text{rel}}$ independently $n_e$ times, each with temperature sampling. Analogous to node extraction, we keep only those triples $(x,r,y)$ that occur in at least $\tau_e$ runs, improving recall while suppressing hallucinated relations, where $x$ and $y$ denote character entities and $r$ denotes the relation type:
\begin{equation*}
\resizebox{0.95\linewidth}{!}{$
  \mathcal{E}_{\text{raw}}
  = \bigl\{(x,r,y)\mid
    \textstyle\sum_{i=1}^{n_e}
    \mathbb{I}\!\bigl((x,r,y)\in\mathcal{F}_{\text{rel}}^{(i)}(\mathcal{D},\mathcal{V})\bigr)
    \ge \tau_e
  \bigr\}.
$}
\end{equation*}

\paragraph{Symbolic Reasoning}
LLMs are essentially sequence-based models and often lack a holistic view of the relational graph within a given text. As a result, they frequently miss links when constructing the predicted relation graph~\citep{reversal-curse2023}. Broadly, two types of missing links can occur: (1) \textbf{Incomplete relations}, where a unidirectional relationship such as "A to B" is present, but the reverse "B to A" is absent; and (2) \textbf{Incomplete structures}, where both "A to B" and "B to A" are missing. Incomplete structures can further be categorised into two subtypes. First, when relationships like "A to B" and "A to C" exist, an inferable relation such as "B to C" (and its reverse "C to B") should be logically deduced, but is missing. Second, although both "A to B" and "B to A" exist, the underlying semantic relations, such as hypernymy or hyponymy, are not identified.
\begin{figure}[t!]
    \centering
    \includegraphics[width=\linewidth]{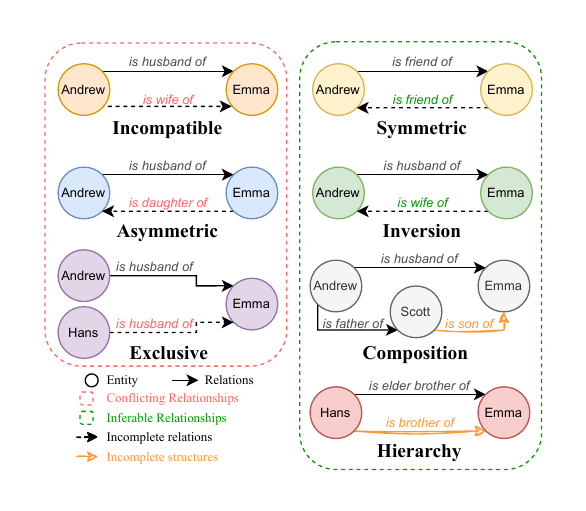}
    \caption{Patterns of interpersonal relationships.}
    \label{fig:relation-pattern}
    \vskip -0.1in
\end{figure}

Illustrative examples are provided in Figure~\ref{fig:relation-pattern}. To address these issues, we propose auto-completing four categories of inferable relationships, enabling existing triples to entail their logically implied but previously missing counterparts.

(1) \emph{Symmetry}: if a relationship $r_1$ holds between $x$ and $y$, then the same relationship holds between $y$ and $x$ ($r_1(x,y) \implies r_1(y,x)$). 

(2) \emph{Inversion}: if a relationship $r_1$ is present between $x$ and $y$, then a different relationship $r_2$ exists between $y$ and $x$ ($r_1(x,y) \implies r_2(y,x)$). 

(3) \emph{Composition}: if a relationship $r_1$ connects $x$ and $y$, and a relationship $r_2$ links $y$ and $z$, then a third relationship $r_3$ can be inferred between $x$ and $z$ ($r_1(x,y) \cap r_2(y,z) \implies r_3(x,z)$). 

(4) \emph{Hierarchy}: Certain relationships can be classified as subtypes or special cases of more general relationships ($r_1(x,y) \subseteq r_2(x,y)$).

We also identify \emph{conflicting relationships}, where the presence of one relation precludes another.

(1) \emph{Incompatible}: The presence of a relationship $r_1$ between individuals $x$ and $y$ is incompatible with the presence of another relationship $r_2$ between the same individuals ($r_1(x,y) \implies \neg r_2(x,y)$). 

(2) \emph{Asymmetric}: if a relationship $r_1$ exists between $x$ and $y$, then the inverse relationship does not hold between $y$ and $x$ ($r_1(x,y) \implies \neg r_2(y,x)$). 

(3) \emph{Exclusive}: if a specific relationship $r_1$ exists between individuals $x$ and $y$, then $x$ cannot have the same relationship with any other individual $z$ ($r_1(x,y) \implies \forall z, \neg r_1(x,z) $).

However, such conflicts often lack clear directionality (i.e., it is unclear which side is incorrect).  
To resolve this, we retrieve relevant context using a retrieval function \(f(r_1, r_2, \mathcal{D})\),  
where \(f\) is implemented via a FAISS-based retriever~\citep{lewis2020retrieval, douze2024faiss} that selects supporting evidence from the source text.  
The retrieved context is then used to construct multiple-choice prompts that help LLMs revise the conflicting relations.

Thus, one of the significant benefits of this annotation and automatic refinement framework is the improved \textbf{completeness}, which means that:

\newtheorem{theorem}{Theorem}

\begin{theorem}
Let $\mathcal{V}$ denote the set of entities, $\mathcal{E}$ the set of relations, and $\mathcal{G}_o = (\mathcal{V}, \mathcal{E}_o)$ the initial relation graph constructed by the LLM. Given a ground-truth relation graph $\mathcal{G}_g = (\mathcal{V}, \mathcal{E}_g)$, there must exist a finite sequence of operations $O_1, O_2, \ldots, O_k$, where each $O_i$ belongs to the set of operations defined in our symbolic reasoning module, such that: $\mathcal{G}_o \xrightarrow{O_1,O_2,...,O_k}\mathcal{G}_g$
always holds.\footnote{The proof is provided in Appendix~\ref{proof_1}.}
\end{theorem}

\paragraph{Interactive Human Verification.}  
During annotation, relations suggested by the model are highlighted in yellow.  Annotators can confirm these suggestions, turning them green. Relations added manually by annotators are also marked in green by default.  For each newly added relation, we automatically perform symbolic reasoning to infer additional triples;  these auto-completed relations are again marked in yellow, pending verification.  If any conflicting relationships are detected during annotation, they are highlighted in red,  with hoverable tooltips showing which existing relation they conflict with. Additionally, when a user selects a relation to confirm or modify, we highlight the corresponding supporting evidence $f(r_1, r_2, \mathcal{D})$ in the original text to assist their decision.

\section{The Relationship Logic Dataset} 
\label{sec:dataset}

\paragraph{LLMs on Logical Reasoning}
\label{sec:logical-reasoning}
We evaluated the performance of LLMs in inferring logical connections among character relationships. The \textbf{Add} task focuses on the open-ended inference of new relationships from existing ones, where multiple correct answers exist and the label space is large. We evaluate model performance using \textit{precision}, \textit{recall}, and \textit{F1-score}. The \textbf{Remove} task, on the other hand, determines whether a pair of relationships conflict with each other, framed as a three-way classification (\textit{Yes}, \textit{No}, \textit{Unsure}); for this, we report \textit{accuracy} and \textit{F1 score}. Table~\ref{tab:llm_performance} reveals that even state-of-the-art models struggle to achieve satisfactory results, highlighting that LLMs often lack a grounded understanding of logical relationships and are limited by training data biases. This underperformance underscores the need for more reliable annotation strategies and robust human-in-the-loop frameworks to support consistent and contextually appropriate reasoning in complex narrative texts.

\begin{table} [h]
    \centering
    \resizebox{\columnwidth}{!}{
    \begin{tabular}{l c c c c c}
    \toprule
    \textbf{Model} & \multicolumn{3}{c}{\textbf{Add}} & \multicolumn{2}{c}{\textbf{Remove}} \\
    \cmidrule(lr){2-4} \cmidrule(lr){5-6}
    & Precision & Recall & F1 & Accuracy & F1 \\
    \midrule
    GPT-4.1      & 71.9 & 55.0 & 62.3 & 57.3 & 42.8 \\
    GPT-4o-mini  & 63.7 & 48.8 & 55.3 & 30.6 & 36.7 \\
    Qwen2.5-32B-Ins  & 64.7 & 43.8 & 52.2 & 31.6 & 34.7 \\
    \bottomrule
    \end{tabular}
    }
    \caption{Performance comparison of LLMs across logical relationship tasks. }
    \label{tab:llm_performance}
    \vskip -0.1in
\end{table}

\paragraph{Expert Annotation of Logical Relationships}
To assess LLM performance in logical reasoning over character relationships, two trained experts annotated conflicts and antisymmetric interactions across 160 character pairs, resulting in 51k derived logical relations. Inter-annotator agreement, measured by Cohen’s Kappa~\cite{cohen1960coefficient}, was 0.832, indicating substantial consistency. All disagreements were resolved through discussion.

\paragraph{Corpus Collection and Relation Extraction}
To facilitate the evaluation of model understanding across diverse textual domains, we curated a corpus of 19 narrative texts spanning four categories: historical accounts, news reports, biographical narratives, and fictional stories. From these, we extracted 1,398 relation triplets among characters (see Table~\ref{tab:dataset_statistics} for details).

\paragraph{Usability Assessment by Non-expert Annotators}
To further evaluate the annotation system, we recruited 10 non-expert annotators to assess its usability and user experience. Detailed annotation instructions and additional statistics are provided in Appendix~\ref{sec:annotation}.


\section{Evaluation}
\subsection{Experimental Setup}

We conducted experiments with OpenAI’s GPT-4.1 (gpt-4.1-2025-04-14), GPT-4o-mini (gpt-4o-mini-2024-07-18), and Qwen2.5-32B-Instruct\footnote{\url{https://huggingface.co/Qwen/Qwen2.5-32B-Instruct}}. For retrieval-augmented generation (RAG), we built the vector library using the text-embedding-3-small model via the OpenAI API\footnote{\url{https://platform.openai.com/docs/overview}}.
We compare \textit{SymbolicThought} with the following baselines: 
(1) \textbf{Direct Prompting}, where the model is directly prompted with instructions; 
(2) \textbf{Self-Consistency} \citep{wang2023selfconsistency}, which samples multiple outputs and selects the final answer via majority voting; 
(3) \textbf{Self-Reflection} \citep{chain-of-thoughts-wei2022}, where the model generates initial predictions and then refines them through self-critique; 
(4) \textbf{Human Annotation}, representing fully manual labeling. 

\subsection{Experimental Results}

\paragraph{Character Extraction}
As shown in Table~\ref{tab:acquisition}, all models benefit from the self-consistency, we adopt for character extraction, with consistent F1-score gains over direct prompting.  
\begin{table}[!htbp]
\centering
\resizebox{\columnwidth}{!}{
\begin{tabular}{l c c c c}
\toprule
\textbf{Model} & \textbf{Method} & \textbf{Precision} & \textbf{Recall} & \textbf{F1} \\
\midrule
\multirow{2}{*}{\textbf{GPT-4.1}} 
& Prompt & 76.9 & 83.1 & 79.8 \\
& Self Consistency & \textbf{77.1} & \textbf{86.7} & \textbf{81.6} \\
\midrule
\multirow{2}{*}{\textbf{GPT-4o-mini}} 
& Prompt & 82.1 & 75.8 & 78.8 \\
& Self Consistency & \textbf{83.1} & \textbf{77.4} & \textbf{80.2} \\
\midrule
\multirow{2}{*}{\textbf{Qwen2.5-32B-Ins}} 
& Prompt & 81.0 & 79.0 & 80.0 \\
& Self Consistency & \textbf{83.5} & \textbf{79.4} & \textbf{81.4} \\
\bottomrule
\end{tabular}
}
\caption{Performance comparison of models and methods on character extraction.}
\label{tab:acquisition}
\vspace{-0.1em}
\end{table}

\begin{table*}[h!]
\centering
\resizebox{0.95\linewidth}{!}{
\begin{tabular}{lccccccccc}
\toprule
\multirow{2}{*}{\textbf{Method}} 
& \multicolumn{3}{c}{GPT-4.1}  
& \multicolumn{3}{c}{GPT-4o-mini}  
& \multicolumn{3}{c}{Qwen2.5-32B-Ins}  \\
\cmidrule(lr){2-4} \cmidrule(lr){5-7} \cmidrule(lr){8-10}
& Precision & Recall & F1-score 
& Precision & Recall & F1-score 
& Precision & Recall & F1-score  \\
\midrule
Prompting                        & 30.9 & 36.4 & 33.4 & 20.6 & 6.5 & 9.9 & 20.9 & 11.5 & 14.8 \\
Self-Consistency        & 29.4 & 35.0 & 32.0 & 27.2 & 8.1 & 12.5 & 23.9 & 13.1 & 16.9 \\
Self-Reflection                  & 24.5 & 26.6 & 25.5 & 22.4 & 7.8 & 11.5 & 20.8 & 12.8 & 15.8 \\
\textit{SymbolicThought}         & \textbf{35.1} & \textbf{41.2} & \textbf{37.9} & \textbf{32.2} & \textbf{13.2} & \textbf{18.8} & \textbf{27.7} & \textbf{18.9} & \textbf{22.5}
\\
\midrule
\end{tabular}}
\caption{Performance comparison of relation extraction using various methods.}
\label{tab:main-results}
\end{table*}

\paragraph{Relation Extraction.}
As Table~\ref{tab:main-results} shows, relation extraction from narrative texts remains a highly challenging task for existing LLMs, with all baseline methods yielding relatively low F1 scores, reflecting
fragmented reasoning and weak cross‐sentence handling. Typical conflicts and further error analysis are in
Appendix~\ref{sec:case-studies}. In contrast, our proposed \textit{SymbolicThought} method, grounded in symbolic reasoning, consistently outperforms all baselines across models.

\paragraph{Human vs. System Performance.}
For evaluation, we merge system-generated and human annotations to form the final ground truth. As shown in Table~\ref{tab:llama-comparison}, \ModelName\ achieves substantial improvements over human annotators across all narrative genres, with over 25\% higher recall and up to 40\% less annotation time. These results confirm the effectiveness of our human-in-the-loop design in streamlining annotation and supporting nuanced social judgments.

\subsection{Annotator Feedback Results}
\begin{figure*}[th!]
    \centering
    \includegraphics[width= \linewidth]{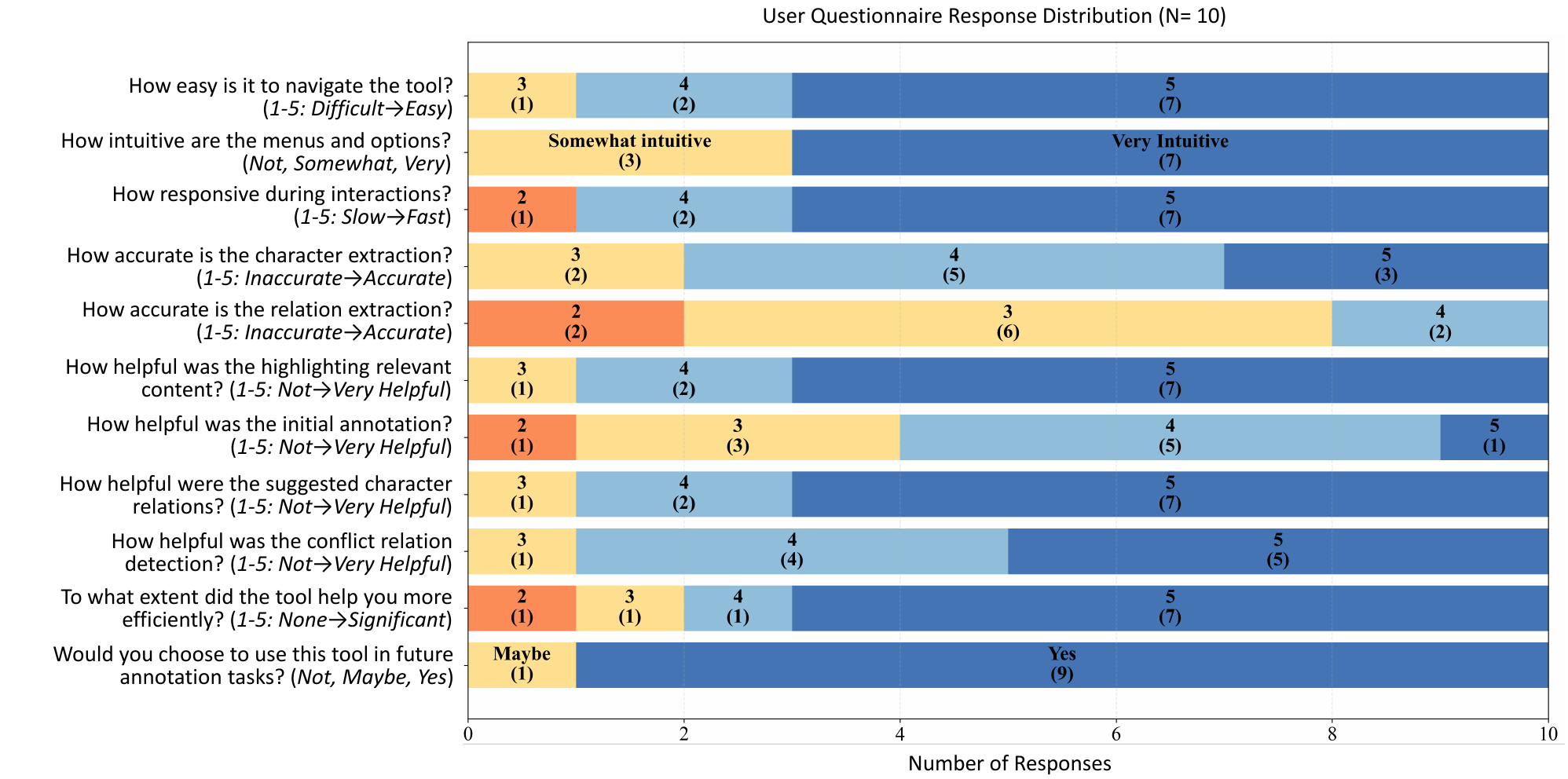}
    \caption{\small{Full Questionnaire Response Distribution. Shades of blue denote more favourable evaluations of the tool, while shades of red indicate dissatisfaction. Grey segments represent neutral factual responses that are not direct indicators of user sentiment toward the tool.}}
    \label{fig:questionnaire}
    \vspace{-0.1em}
\end{figure*}

To evaluate the usability and effectiveness of \textit{SymbolicThought}, we conducted a user study involving 10 participants. Each participant interacted with the system and completed a structured questionnaire (Figure~\ref{fig:questionnaire}), which covered topics such as interface usability, system performance, feature usefulness, and overall user experience. The complete questionnaire is provided in Appendix~\ref{sec:questionnaire}.

As shown in Figure~\ref{fig:questionnaire}, users found the tool highly intuitive and responsive. The majority of users considered the system-generated initial annotations and automatic relation suggestions helpful in reducing their workload. Notably, features such as text highlighting and conflict detection received consistently high usefulness ratings, especially the automatically suggested character relations. While the accuracy of character extraction was generally well received, users reported slightly lower confidence in relation extraction, suggesting opportunities for further refinement. Overall, 9 out of 10 participants indicated they would choose to use the tool in future annotation tasks, confirming the practical value of \textit{SymbolicThought} in supporting efficient and accurate narrative analysis.

\subsection{Correlation Analysis: Annotation Complexity vs. Small World Index}
\begin{figure}[t!]
    \centering
    \includegraphics[width=0.9 \linewidth]{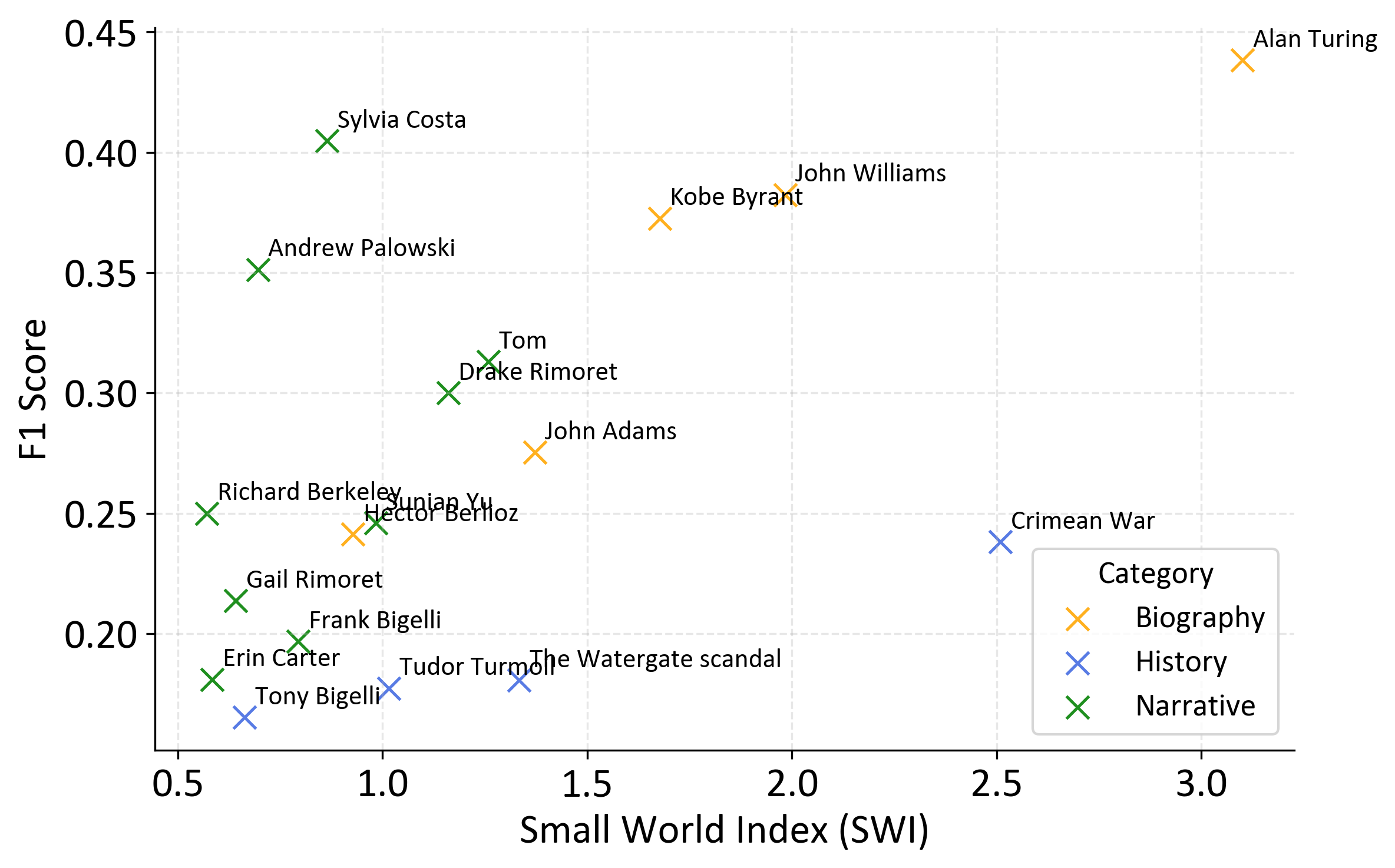}
    \caption{F1 score vs. Small World Index. Each point represents a document, colored by its category.}
    \label{fig:correlation_swi_f1}
    \vskip -0.1in
\end{figure}

Figure~\ref{fig:correlation_swi_f1} plots annotation performance (F1) against the
\textit{Small-World Index}~(SWI) \citep{telesford2017small}, which is a widely used index from social science research to measure the compactness of a given graph of social relationships. 
A larger SWI therefore indicates an entity graph that is both \emph{locally dense} (many closed triads) and \emph{globally efficient} (short paths), i.e.\ more ``small-world-like''.

\paragraph{Cross‐category differences.}
Biography documents cluster in the upper-right portion of the plot, yielding significantly higher F1 scores than Novels and History texts. This suggests that biographies possess more cohesive entity graphs, making them easier to annotate. Novels occupy the middle range, whereas History texts lie predominantly in the lower-left, reflecting their sparser, less structured networks and hence greater annotation difficulty.

\paragraph{Within-category trend.}
Across all three categories, F1 increases monotonically with SWI, indicating that richer small-world structure reduces annotation complexity. In practice, SWI thus serves as a reliable proxy for predicting how much human effort a document will require during annotation.

\begin{table}[!htbp]
    \centering
    \vspace{0.1in}
    \resizebox{\linewidth}{!}{
        \begin{tabular}{lcccc}
            \toprule[1pt]
            Category & Method & History  & Biography & Narrative \\
            \hline
            \multirow{2}{*}{Recall (\%)} & Human& 57.3 & 67.3& 63.4 \\
            & \ModelName &\textbf{85.6} & \textbf{91.4} & \textbf{89.1} \\
            \hline
            \multirow{2}{*}{Avg Time (min)} & Human &163.4 &  87.2& 102.5  \\
            & \ModelName & \textbf{118.3 }& \textbf{45.5} & \textbf{74.7} \\
            \bottomrule[1pt]
        \end{tabular}
    }
\caption{Comparison of recall and annotation time between human and our system across different genres.}
    \label{tab:llama-comparison}
\end{table}

\section{Conclusion}
We introduced \textit{SymbolicThought}, a comprehensive human-in-the-loop framework that integrates LLM extraction with symbolic reasoning for more accurate and reliable character relationship annotation. Our system effectively bridges the gap between model predictions and human expectations, ensuring logical consistency, interpretability, and transparency in the analysis of complex narrative texts.

\section*{Acknowledgments}
This work was supported in part by the UK Engineering and Physical Sciences Research Council (EPSRC) through a Turing AI Fellowship (grant no. EP/V020579/1, EP/V020579/2) and a New Horizons grant (grant no. EP/X019063/1), and KCL’s Impact Acceleration Account (grant no. EP/X525571/1). A PhD studentship from the Chinese Scholarship Council funds Qinglin Zhu. The authors also acknowledge the use of the King’s Computational Research, Engineering, and Technology Environment (CREATE) at King’s College London. 

\bibliography{custom}

\appendix

\section{Annotation Details}
\label{sec:annotation}
For \textit{Symmetry \& Inversion} relationships, the label space encompassed the complete set of relationships, $R$. For \textit{Asymmetric} and \textit{Incompatible} relationships, the label space consisted of \{``Yes'', ``No'', ``Unsure''\}. The human annotators used the simpler label space \{``Yes'', ``No''\}. Because some relationships were challenging to classify, we labelled all disagreements between annotators as ``Unsure''.
For \textit{Composition} relationships, the label space spans all 160 predefined character relationship categories.

In addition to the expert-labeled logical relationships, we also recruited 10 non-expert annotators to test the usability of our annotation system. These annotators were given standardised training materials, identical to the video-based tutorial accompanying this paper. This setup allows us to evaluate the accessibility and efficiency of the system when used by general users rather than domain experts. 
Each expert annotator completed approximately 40 hours of work and was compensated at a rate of \$31.92 per hour. 

\begin{table}[h]
\centering
\resizebox{1 \columnwidth}{!}{
\begin{tabular}{@{}lrrr@{}}
    \toprule
    \textbf{Dataset}  & \textbf{Historical} & \textbf{Biography} & \textbf{Narrative} \\
    \midrule
     \#Narratives &  4 &  5 & 10  \\
    Avg. \#Character  & 16.4  &  24.2 & 9.0  \\ 
    \#Relationships & 85.0  &  130.0 &  48.7 \\
    Avg. \#Token & 1246.9  & 737.1  &  3467.3 \\
    \bottomrule
\end{tabular}
}

\caption{Dataset Statistics.}
\label{tab:dataset_statistics}
\end{table}

\section{User Questionnaire}
\label{sec:questionnaire}

To better understand user needs and evaluate the usability and effectiveness of \textbf{SymbolicThought}, we designed a structured questionnaire targeting key aspects of the user experience. The survey covers four major dimensions: interface design, system performance, feature usefulness, and overall feedback. The full questionnaire is provided below.

\vspace{1em}

\begin{tcolorbox}[title=Section 1: Interface Design, colback=gray!5, colframe=gray!40!black, breakable, fonttitle=\small, fontupper=\small]
\textbf{Q1.} How easy is it to navigate the tool? \\
(1 = Very Difficult, 5 = Very Easy)

\vspace{0.8em}
\textbf{Q2.} How intuitive are the menus and options? \\
- Very intuitive \\
- Somewhat intuitive \\
- Not intuitive
\end{tcolorbox}

\begin{tcolorbox}[title=Section 2: System Performance, colback=gray!5, colframe=gray!40!black, breakable, fonttitle=\small, fontupper=\small]
\textbf{Q3.} How responsive is the tool during interactions? \\
(1 = Very Slow, 5 = Very Fast)

\vspace{0.8em}
\textbf{Q4.} How accurate is the character extraction provided by the tool? \\
(1 = Very Inaccurate, 5 = Very Accurate)

\vspace{0.8em}
\textbf{Q5.} How accurate is the relation extraction provided by the tool? \\
(1 = Very Inaccurate, 5 = Very Accurate)

\vspace{0.8em}
\textbf{Q6.} What is the maximum acceptable wait time for character and relation extraction? \\
- Less than 1 minute \\
- Less than 5 minutes \\
- Depends on the length of the narrative \\
- Less than the time it would take me to do it manually \\
- As long as the quality is high
\end{tcolorbox}

\begin{tcolorbox}[title=Section 3: Feature Usefulness, colback=gray!5, colframe=gray!40!black, breakable, fonttitle=\small, fontupper=\small]
\textbf{Q7.} How helpful was the system-generated initial annotation (i.e., the AI-suggested pre-filled characters and relation graph) in reducing your workload? \\
(1 = Not Helpful, 5 = Very Helpful)

\vspace{0.8em}
\textbf{Q8.} How helpful was the text highlighting of relevant content? \\
(1 = Not Helpful, 5 = Very Helpful)

\vspace{0.8em}
\textbf{Q9.} How helpful were the automatically suggested character relations? \\
(1 = Not Helpful, 5 = Very Helpful)

\vspace{0.8em}
\textbf{Q10.} How helpful was the conflict relation detection/reminder? \\
(1 = Not Helpful, 5 = Very Helpful)
\end{tcolorbox}

\begin{tcolorbox}[title=Section 4: Overall Feedback, colback=gray!5, colframe=gray!40!black, breakable, fonttitle=\small, fontupper=\small]
\textbf{Q11.} To what extent did the tool help you complete the annotation task more efficiently? \\
(1 = Not at All, 5 = Significantly)

\vspace{0.8em}
\textbf{Q12.} Would you choose to use this tool in future annotation tasks? \\
- Yes \\
- Maybe \\
- No

\end{tcolorbox}

\section{Completeness of operations}
\label{proof_1}

\newtheorem{lemma}{Lemma}
\newtheorem{corollary}{Corollary}

To prove \textbf{Theorem 1}, we first consider a scenario with the simplest assumption. 

\begin{lemma}
    \textbf{Theorem 1} is true when there is only one type of relation defined in $\mathcal{E}$.
\end{lemma}

\noindent \textbf{Proof.} When there is only one type of relation defined in $\mathcal{E}$, the graph $\mathcal{G} = (\mathcal{V}, \mathcal{E})$ can be represented as a Boolean adjacency matrix. Specifically, let $x_i$ and $x_j$ be the $i$-th and $j$-th entities, respectively. If there exists a relation between them, then the corresponding matrix entry is $\mathcal{G}_{ij} = 1$; otherwise, it is 0.

Both $\mathcal{G}_o$ (the output graph from the LLM) and $\mathcal{G}_g$ (the ground-truth graph) are Boolean matrices. By assumption, $\mathcal{G}_o$ contains a subset of the non-zero entries in $\mathcal{G}_g$. Therefore, there exists a Boolean matrix $\mathcal{G}_a$ such that:

\begin{equation}
    \mathcal{G}_o + \mathcal{G}_a = \mathcal{G}_g   \nonumber 
\end{equation}

where the addition is element-wise Boolean OR.

Since the number of entities is finite, the matrices $\mathcal{G}_o$, $\mathcal{G}_a$, and $\mathcal{G}_g$ each contain a finite number of non-zero (i.e., 1) entries.

Under the assumption of a single relation type, the reasoning operations are limited to two forms:

\begin{itemize}
    \item \noindent  \textbf{T1 (Symmetry):} If $\mathcal{G}_{ij} = 1$, then infer $\mathcal{G}_{ji} = 1$.
    \item  \noindent \textbf{T2 (Transitivity):} If $\mathcal{G}_{ij} = 1$ and $\mathcal{G}_{jk} = 1$, then infer $\mathcal{G}_{ik} = 1$.
\end{itemize}

Each operation corresponds to the addition of a single 1-entry in the matrix. Thus, we can construct a finite sequence of such operations to transform $\mathcal{G}_o$ into $\mathcal{G}_g$ in the following steps:

\begin{itemize}
    \item \textbf{Symmetry Completion:} For each pair $(i, j)$ where $\mathcal{G}_{ij} = 1$ appears in $\mathcal{G}_o$ but $\mathcal{G}_{ji} = 1$ only exists in $\mathcal{G}_g$, we apply a sequence of \textbf{T1} operations. Let the number of such pairs be $s$, and denote the operations as $\sum_{i=1}^{s} O_i$.
    \item \textbf{Transitive Completion:} For pairs $(i, j)$ such that $\mathcal{G}_{ij} = 1$ appears in $\mathcal{G}_g$ but not in $\mathcal{G}_o$, and where $i < j$, we apply \textbf{T2} operations to infer these missing links. Let the number of such inferences be $t$, and denote the operations as $\sum_{i=s+1}^{s+t} O_i$.
    \item \textbf{Symmetry for New Edges:} Some of the newly added relations in Step 2 may require symmetric counterparts to satisfy $\mathcal{G}_g$. We again apply \textbf{T1} operations for these cases. Let the number of these be $k - (s + t)$, and denote them as $\sum_{i=s+t+1}^{k} O_i$.
\end{itemize}

By composing these operations, we construct a finite sequence ${O_1, O_2, \ldots, O_k}$ such that:

\begin{equation}
    \mathcal{G}_o \xrightarrow{O_1,O_2,...,O_k}\mathcal{G}_g  \nonumber 
\end{equation}

\hfill$\blacksquare$

Then, we can extend the \textbf{Lemma 1} in to a more complex scenario which contains two types of relations.

\begin{corollary}
    \textbf{Theorem 1} is true when there are two types of relations defined in $\mathcal{E}$.
\end{corollary}

\noindent \textbf{Proof.} When two relation types are defined in $\mathcal{E}$, the relation graph can be represented by two corresponding Boolean matrices, each encoding one type of relation. Extending the reasoning in \textbf{Lemma 1}, each type of completion (symmetry and transitivity) now applies independently to each relation type. As a result, there are:

\begin{itemize}
    \item 4 possible cases for symmetry completion (2 relation types × direction),
    \item 8 possible cases for transitive completion (combinations across 2 relation types).
\end{itemize}

In addition, when there are more than one type of relation, we need to consider an additional type of completion, referred to as \textbf{copy completion}. This occurs when the presence of a relation of one type implies the presence of a corresponding relation of another type. There are 2 possible cases of copy completion, depending on the direction and pairing of relation types.

Given that the number of entities is finite, and the total number of possible relational triples is bounded, the number of necessary completions remains finite. Thus, it is still possible to construct a finite sequence of operations $O_1, O_2, \ldots, O_k$ such that:

\begin{equation}
    \mathcal{G}_o \xrightarrow{O_1,O_2,...,O_k}\mathcal{G}_g  \nonumber 
\end{equation}

\hfill$\blacksquare$

Based on the results of \textbf{Lemma 1} and \textbf{Corollary 1}, the proof of \textbf{Theorem 1} follows directly. For multiple relation types, we can first sort the relations by their frequency of occurrence and then traverse the entire graph. By incrementally editing the relations one by one, we construct a finite sequence of operations holds for all relation types.

\section{Case Studies}
\label{sec:case-studies}
\begin{figure*}[th!]
    \centering
    \includegraphics[width=\linewidth]{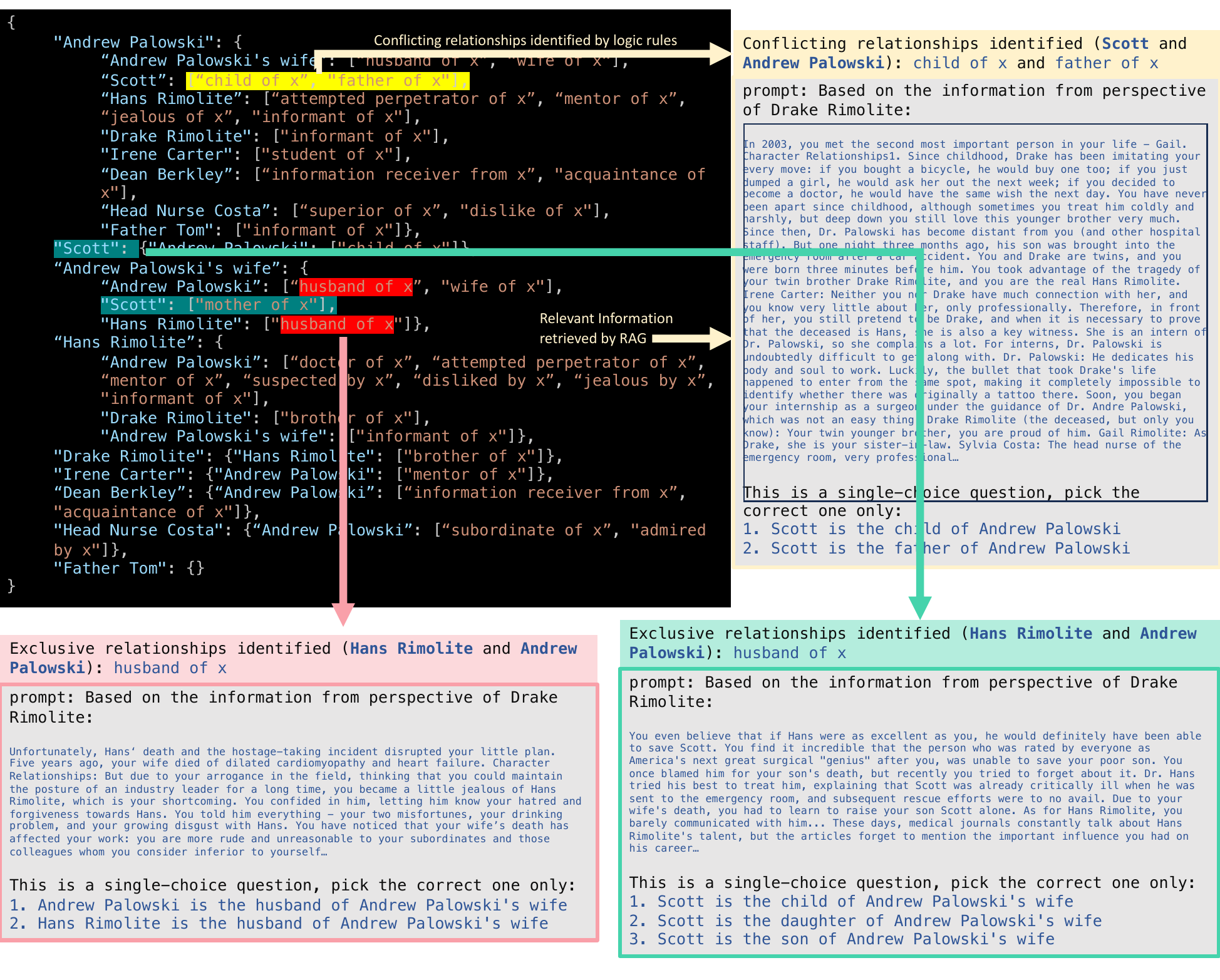}
    \caption{A demonstration of our relationship refinement pipeline. Starting from rough predictions (top left figure), we leverage the logic relationships to identify erroneous relationship edges. Depending on the type of logical conflict, we support LLMs with context retrieved via RAG and a multiple-choice prompt to refine the relationships.}
    \label{fig:relation_refinement}
\end{figure*}
As shown in Figure~\ref{fig:relation_refinement}, the initial relation graph extracted by LLMs often contains conflicting information. These include simple incompatibilities (e.g., ``child of'' vs. ``father of'') highlighted in yellow, as well as more subtle exclusive contradictions (e.g., multiple conflicting ``husband of'' links) highlighted in red. By defining logical constraints over relation types, such as mutual exclusivity or inverse relation, we can automatically identify these inconsistencies in the graph.

To resolve such conflicts, we retrieve supporting evidence from the original narrative using a RAG module. For example, in the top right of Figure~\ref{fig:relation_refinement}, a logical conflict is detected between two claims: ``Scott is the child of Andrew Palowski'' vs. ``Scott is the father of Andrew Palowski.'' Relevant context is retrieved from the text and presented to the LLM as part of a structured single-choice question. The LLM then selects the logically consistent option based on grounded evidence.

Similarly, in the lower part of the figure, we show an example of exclusive relationship refinement involving two candidates for the husband of Andrew Palowski’s wife. The tool retrieves narrative evidence from multiple perspectives and uses contrastive prompting to resolve the ambiguity. In both examples, the system leverages the symbolic logic layer to surface conflicts and guide the LLM’s reasoning, while maintaining human interpretability through structured prompts and explicit evidence grounding.

This refinement pipeline allows \textit{SymbolicThought} to iteratively improve the coherence and correctness of the extracted relation graph, addressing one of the key limitations of LLMs, namely, inconsistent or contradictory relational inference in complex narrative contexts.

\subsection{Error Analysis}
Below, we categorise and analyse the common failure modes observed during our experiments.

\paragraph{What kind of relationships do LLMs miss?}
LLMs often miss relationships that require some inference, such as \textit{``sister-in-law''}, \textit{``brother-in-law''}, \textit{``father-in-law''}, etc. These relationships are often not explicitly stated in the original text. For example, A is B's wife, and B is C's child, so A should be C's daughter-in-law. In such cases, LLMs often fail to successfully capture this information.
\paragraph{What kind of hallucinations do LLMs generate?}
There are differences in how humans and machines understand relationships. For example, consider the following text: \textit{``When you were in dire straits, you often confided in Father Tom, who visited Brighton Hospital. You told him your misfortunes, your drinking problem. Tom tried his best to comfort you, and on his advice, you participated in a treatment program led by a psychiatrist...''}. In this context, human annotators believe that the priest helped Andrew, so he is a helper and a saviour. However, LLMs perceive him as an information receiver from Andrew. While this interpretation is not entirely incorrect, it is a more superficial relationship. Such differences in understanding lead to many noisy labels generated by LLMs.

\paragraph{What kind of logic mistakes do LLMs make?}
One issue is that LLMs often struggle to identify the direction of relationships. We tried providing few-shot examples and instructions, but these approaches did not yield significant improvements. Another issue is that when we identify a conflicting relationship and ask the model to reflect on which is correct, it often answers both incorrectly, displaying strong biases. To address this, we adopt a multiple-choice format that forces the model to choose the most likely option.

\end{document}